\documentclass[conference]{IEEEtran}
\IEEEoverridecommandlockouts
\usepackage{cite}
\usepackage{amsmath,amssymb,amsfonts}
\usepackage{algorithmic}
\usepackage{graphicx}
\usepackage{textcomp}
\usepackage{xcolor}
\usepackage{booktabs}
\usepackage{algorithm}
\usepackage{amsfonts}
\usepackage{multirow}
\usepackage{multicol}
\def\BibTeX{{\rm B\kern-.05em{\sc i\kern-.025em b}\kern-.08em
    T\kern-.1667em\lower.7ex\hbox{E}\kern-.125emX}}
\begin{document}

\title{SkinCLIP-VL: Consistency-Aware Vision-Language Learning for Multimodal Skin Cancer Diagnosis}

\author{
    Zhixiang Lu\textsuperscript{1,2}, 
    Shijie Xu\textsuperscript{1}, 
    Kaicheng Yan\textsuperscript{1}, 
    Xuyue Cai\textsuperscript{1}, 
    Chong Zhang\textsuperscript{1}, \\
    Yulong Li\textsuperscript{1}, 
    Angelos Stefanidis\textsuperscript{1}, 
    Anh Nguyen\textsuperscript{2$\dagger$},
    Jionglong Su\textsuperscript{1$\dagger$}\thanks{$\dagger$ Corresponding authors.} \\
    \textsuperscript{1}\textit{Xi'an Jiaotong-Liverpool University, China}  \space \space
    \textsuperscript{2}\textit{University of Liverpool, United Kingdom} \\
    \texttt{jionglong.su@xjtlu.edu.cn, anh.nguyen@liverpool.ac.uk}
}
\maketitle

\begin{abstract}
The deployment of vision-language models (VLMs) in dermatology is hindered by the trilemma of high computational costs, extreme data scarcity, and the ``black-box" nature of deep learning. To address these challenges, we present SkinCLIP-VL, a resource-efficient framework that adapts foundation models for trustworthy skin cancer diagnosis. Adopting a frozen perception, adaptive reasoning paradigm, we integrate a frozen CLIP encoder with a lightweight, quantized Qwen2.5-VL via low-rank adaptation (LoRA). To strictly align visual regions with clinical semantics under long-tailed distributions, we propose the Consistency-aware Focal Alignment (CFA) Loss. This objective synergizes focal re-weighting, semantic alignment, and calibration. On ISIC and Derm7pt benchmarks, SkinCLIP-VL surpasses 13B-parameter baselines by 4.3–6.2\% in accuracy with 43\% fewer parameters. Crucially, blinded expert evaluation and out-of-distribution testing confirm that our visually grounded rationales significantly enhance clinical trust compared to traditional saliency maps.

\end{abstract}

\begin{IEEEkeywords}
Skin Cancer Diagnosis, Vision-Language Models, Multimodal Learning, Medical AI, Foundation Model    
\end{IEEEkeywords}

\section{Introduction}

The clinical diagnosis of cutaneous malignancies involves a complex cognitive process that synthesizes heterogeneous multimedia data. Dermatologists rely on the interplay between high-resolution dermoscopic imagery (visual modality) and structured patient metadata, such as anatomical site and clinical history (textual modality) \cite{DeepGBTB2026}. While the early detection of melanoma significantly improves survival rates \cite{Esteva2017Dermatologist}, the efficacy of current diagnostic workflows is constrained by the scarcity of specialized expertise, creating a bottleneck in global healthcare systems \cite{Rotemberg2021Patient}. This disparity has catalyzed the demand for automated Clinical Decision Support Systems (CDSS) capable of processing multimodal medical data.

Recent advancements in foundation models (FMs), particularly vision-language models (VLMs) like CLIP \cite{Radford2021Learning} and large language models (LLMs), have demonstrated remarkable capabilities in general multimodal understanding. However, adapting these billion-parameter models to the specialized domain of skin cancer diagnosis faces three systemic challenges. \textbf{First, the Data-Efficiency vs. Scale Paradox.} Training or fully fine-tuning large-scale FMs requires vast annotated datasets. Conversely, medical benchmarks like ISIC are characteristically data-scarce and exhibit a severe long-tailed distribution dominated by benign lesions \cite{Cassidy2022Analysis}. Direct training on such imbalanced data often leads to biased representations, where models overfit to majority classes while failing to generalize to critical yet rare malignancies. \textbf{Second, the Challenge of Heterogeneous Semantic Alignment.} Effective diagnosis demands the seamless integration of pixel-level visual features with high-level clinical semantics. Existing approaches often rely on shallow fusion mechanisms that treat modalities in isolation, failing to bridge the ``semantic gap'' between raw visual signals and medical terminology. This misalignment limits the model's ability to leverage cross-modal correlations for precision diagnosis. \textbf{Third, the ``Black-Box'' Trust Deficit.} In high-stakes healthcare environments, a system that outputs high-confidence predictions without interpretable reasoning is clinically unactionable \cite{Rudin2019Stop}. To ensure seamless integration into clinical workflows, multimedia systems must provide verifiable evidence, specifically, diagnostic rationales that are visually grounded in the lesion regions \cite{Zhang2023BiomedGPT}.
\begin{figure}[t]
  \centering
   \includegraphics[width=0.9\linewidth]{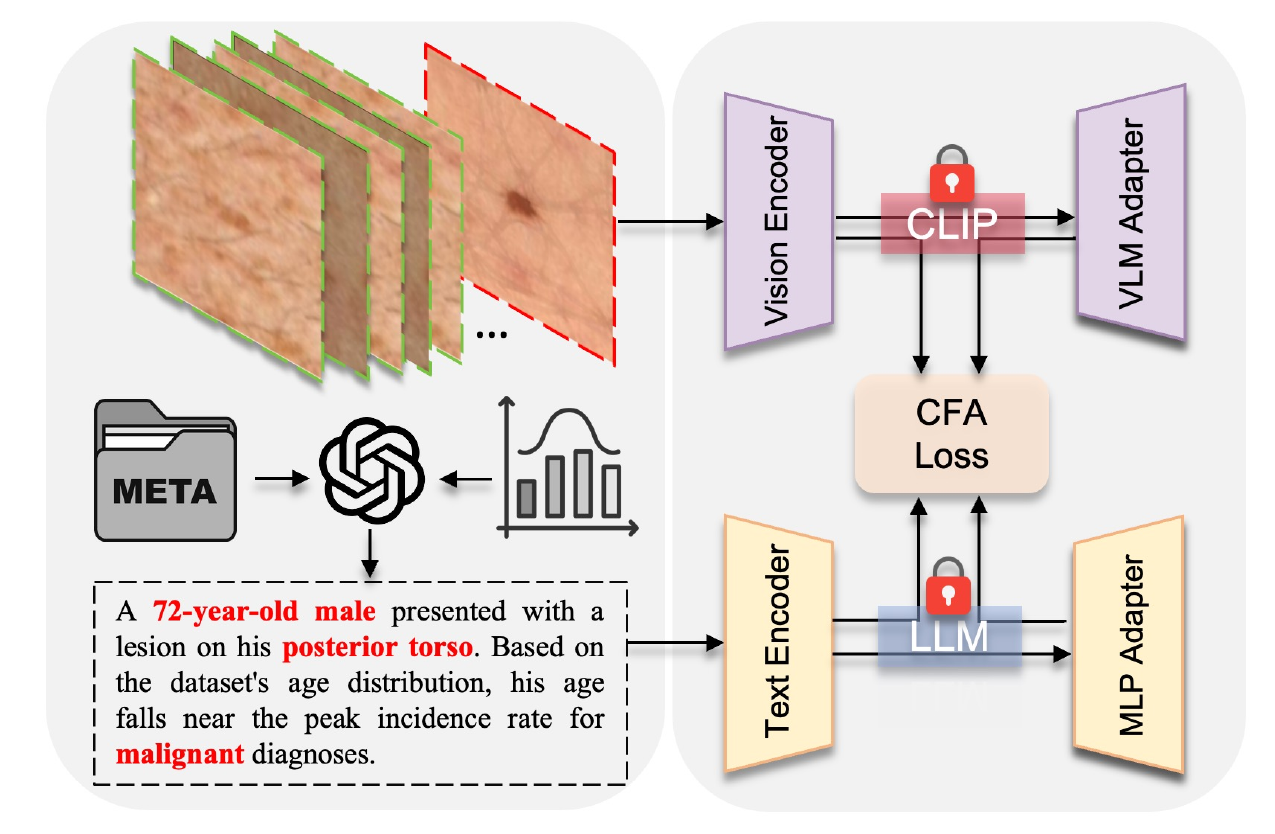}
   \vspace{-1em}
   \caption{Consistency-Aware Focal Alignment Architecture.}
   \label{fig:framework}
   \vspace{-1.5em}
\end{figure}
To bridge the gap between general-purpose FMs and specialized medical requirements, we propose \textbf{SkinCLIP-VL}, a resource-efficient framework for trustworthy multimodal diagnosis. Adopting a ``frozen perception, adaptive reasoning'' paradigm, we integrate a pre-trained CLIP visual encoder with a lightweight Qwen2.5-VL decoder \cite{Bai2023Qwen} via parameter-efficient fine-tuning (PEFT) \cite{Hu2021LoRA}. This architecture leverages robust pre-trained priors while minimizing computational overhead, achieving effective domain adaptation with significantly reduced labeled data. Critically, we introduce the \textbf{Consistency-aware Focal Alignment (CFA) Loss} to orchestrate this multimodal interaction. This unified objective simultaneously addresses: (1) class imbalance via focal re-weighting; (2) cross-modal semantic consistency between visual regions and medical texts; and (3) confidence calibration. Unlike standard black-box classifiers, SkinCLIP-VL generates physician-style rationales grounded in specific image regions, enhancing interpretability. Our main contributions are summarized as follows:

\begin{itemize}
    \item \textbf{Efficient Foundation Model Adaptation:} We present SkinCLIP-VL, a streamlined architecture that reduces trainable parameters by 43\% compared to SOTA baselines, enabling the deployment of advanced VLM capabilities on resource-constrained medical edge devices.

    \item \textbf{Unified Multimodal Optimization:} We introduce the CFA Loss, a novel mechanism that jointly optimizes for imbalanced classification, cross-modal semantic consistency, and predictive reliability.

    \item \textbf{Trustworthy System Validation:} Beyond standard metrics, we validate our system via a blinded expert study ($N=20$), demonstrating that our visually grounded rationales significantly enhance clinician trust compared to traditional saliency maps.
\end{itemize}

\section{Related Work}

The paradigm shift towards FMs, exemplified by CLIP \cite{Radford2021Learning} and recent LLMs, has revolutionized multimedia analysis. While these models possess robust generalization capabilities, adapting them to the medical domain presents significant challenges. Full fine-tuning is not only computationally prohibitive but also susceptible to catastrophic forgetting, effectively erasing the model's pre-trained general knowledge \cite{Jain2024SurveyPEFT}. To mitigate this, PEFT has become the standard for resource-constrained healthcare applications. Techniques such as low-rank adaptation (LoRA) \cite{Hu2021LoRA} and adapter modules allow for effective adaptation by updating only a small fraction of parameters. In the video domain, the frozen perception strategy \cite{Lin2022Frozen} has shown that keeping the visual backbone fixed while training lightweight temporal adapters preserves strong visual priors. However, current medical VLMs often still rely on heavy, trainable visual encoders to bridge the domain gap \cite{Tawde2024DataEfficient}. Our SkinCLIP-VL framework advances this efficiency frontier by strictly freezing the visual backbone and utilizing a quantized LLM decoder, thereby enabling high-performance adaptation on dermatological tasks with minimal trainable parameters and labeled data.

Reliable medical diagnosis requires solving a multi-objective optimization problem involving class imbalance, semantic alignment, and predictive calibration. 
First, dermatological datasets typically exhibit a long-tailed distribution, where benign lesions vastly outnumber malignancies. Standard solutions like Focal Loss \cite{Lin2017Focal} address this by dynamically re-weighting loss contributions to prioritize hard, minority samples. Second, effective Vision-Language Learning hinges on cross-modal alignment. Methods such as MedCLIP \cite{Wang2022MedCLIP} employ InfoNCE-based contrastive learning to align visual and textual feature spaces. However, standard contrastive objectives often treat all negative pairs uniformly, neglecting the nuanced semantic granularity required for ambiguous skin lesions \cite{Wang2021Understanding}. 
Third, confidence calibration is critical for safety, deep models are notoriously overconfident. While techniques like the Dynamically Weighted Balanced (DWB) Loss \cite{DeepGBTB2026} improve calibration error, they are rarely integrated into generative multimodal frameworks. Our work proposes the CFA loss to unify these disjoint objectives: synergizing focal re-weighting, semantic consistency, and calibration into a single differentiable framework.

Clinical adoption of AI systems is contingent upon trust, necessitating transparency beyond scalar probability scores \cite{DeepGBTB2026}. Early Explainable AI (XAI) approaches were predominantly post-hoc, utilizing saliency methods like Grad-CAM \cite{Selvaraju2017GradCAM} to highlight influential regions. While useful, these heatmaps are often low-resolution, noisy, and lack semantic interpretability. The advent of Generative AI has introduced inherent explainability via Chain-of-Thought (CoT) \cite{Wei2022chain} reasoning. However, text-only generation in medicine faces the critical risk of hallucination, where models generate plausible but factually incorrect details \cite{Das2025trustworthy}. The current state-of-the-art (SOTA) is shifting towards visually grounded XAI, which explicitly links generated textual tokens to corresponding image regions \cite{Zhang2023BiomedGPT}. SkinCLIP-VL adopts this direction, generating physician-style rationales where diagnostic terms are mathematically anchored to visual evidence, effectively transforming the ``black box'' into a transparent diagnostic partner.

\begin{figure*}[t]
  \centering
   \includegraphics[width=1\linewidth]{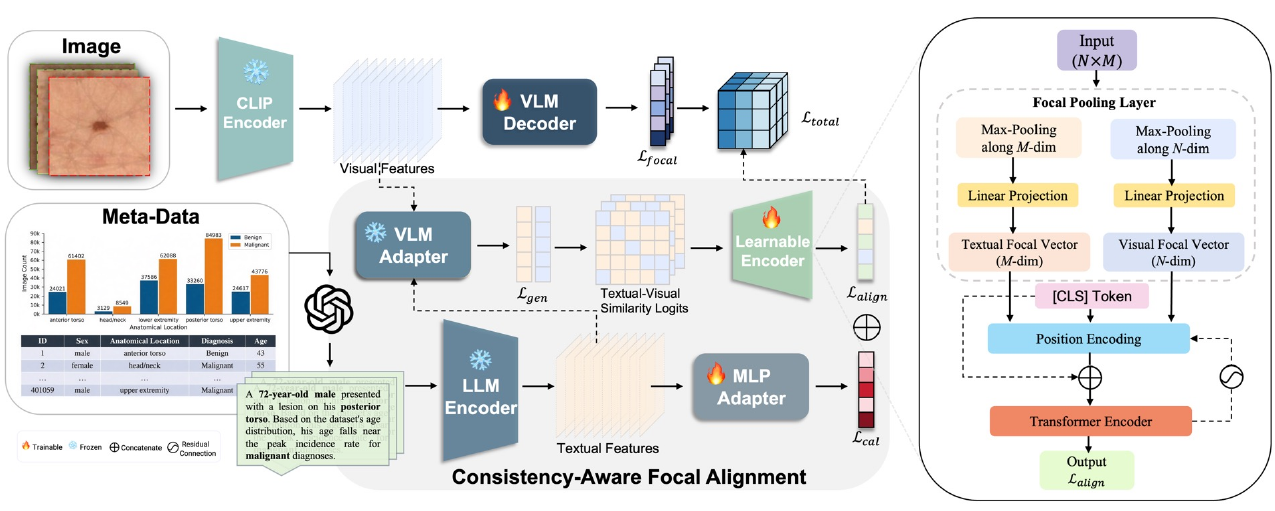}
   \vspace{-2em}
   \caption{The overall framework of SkinCLIP-VL. The architecture consists of three key stages: (1) \textbf{Meta-Data Enhancement}: We leverage GPT-4o to expand tabular meta-data into comprehensive clinical descriptions, providing semantic guidance for the visual branch. (2) \textbf{Parameter-Efficient Encoding}: Instead of full fine-tuning, we employ a frozen CLIP visual encoder and a LoRA-adapted Qwen2.5-VL generative decoder to extract and align visual and textual features. (3) \textbf{Consistency-Aware Focal Alignment (CFA)}: To capture fine-grained correlations, the Focal Pooling Layer aggregates the $N \times M$ interaction map into compact focal vectors. These vectors are fused via a Transformer Encoder under the joint supervision of focal and alignment losses ($\mathcal{L}_{focal}, \mathcal{L}_{align}$).}
   \label{fig:onecol}
   \vspace{-1em}
\end{figure*}

\section{Methodology}

\subsection{Efficient Multimodal Architecture}
The system processes heterogeneous inputs: dermoscopic images ($I$) and structured clinical metadata ($M$) through the integrated modules, as illustrated in Fig. \ref{fig:framework}.

\subsubsection{Frozen Visual Perception ($E_{img}$)}
To harness robust visual priors without incurring the cost of full-scale training, we utilize the Vision Transformer from CLIP \cite{Radford2021Learning} as the visual backbone. To prevent "catastrophic forgetting" of general visual features \cite{Jain2024SurveyPEFT}, the parameters of $E_{img}$ are strictly frozen.
Given an input image $I \in \mathbb{R}^{H \times W \times 3}$, the encoder outputs a sequence of spatial patch embeddings $V = \{v_1, ..., v_N\} \in \mathbb{R}^{N \times d_v}$, where $N$ denotes the number of patches.

\subsubsection{Multimedia Data Bridging}
Clinical diagnosis relies heavily on patient metadata (e.g., age, anatomical site). To align this structured modality with the vision-language latent space, we employ an offline LLM to serialize the tabular metadata $M$ into a natural language description $T_{meta}$ (e.g., \textit{``A lesion on the posterior torso..."}). This converts the heterogeneous fusion problem into a unified vision-language alignment task.

\subsubsection{Focal Pooling \& Feature Projection}
First, a linear projector maps the CLIP embeddings to the LLM's dimension, yielding the input sequence $V_{proj} = \{p_1, ..., p_N\} \in \mathbb{R}^{N \times d_{llm}}$.
To explicitly align visual regions with medical semantics, we parallelly introduce a learnable Focal Pooling Layer. Unlike standard average pooling, this module utilizes a learnable query vector $q_{focal} \in \mathbb{R}^{1 \times d_{llm}}$ to compute an attention-weighted global descriptor $v_{global}$:
\begin{equation}
    \small
    \alpha_i = \text{Softmax}\left(\frac{p_i W_Q (q_{focal} W_K)^T}{\sqrt{d_k}}\right)
\end{equation}
\begin{equation}
    \small
    v_{global} = \sum_{i=1}^{N} \alpha_i \cdot (p_i W_V)
\end{equation}
where $W_Q, W_K, W_V$ are learnable projection matrices. The attention weights $\alpha_i$ effectively highlight clinically salient regions driven by the alignment objective.

\subsubsection{LoRA-Adapted Generative Decoder ($E_{dec}$)}
We employ Qwen2.5-VL-7B \cite{Bai2023Qwen} as the reasoning engine. To achieve parameter efficiency, we apply LoRA \cite{Hu2021LoRA} to the attention weights ($W_q, W_v$) of the decoder, reducing trainable parameters by $\approx 43\%$. The decoder accepts the multimodal sequence $[V_{proj}; T_{meta}]$ and operates in a multi-task manner:
\begin{itemize}
    \item \textbf{Discriminative Head:} A learnable \texttt{[CLS]} token is appended to the input sequence. Its final hidden state $h_{cls}$ is projected to class logits $z \in \mathbb{R}^C$ for classification.
    \item \textbf{Generative Head:} Conditioned on the spatial visual features $V_{proj}$, the model auto-regressively generates the diagnostic rationale $R$.
\end{itemize}

\subsection{Consistency-Aware Focal Alignment (CFA) Loss}
Standard multi-task learning often suffers from "objective conflict" in medical settings \cite{Zhang2023BiomedGPT}. We formulate the training as a synergistic optimization problem via the CFA Loss. Let $\lambda_{1}, \lambda_{2}, \lambda_{3}$ denote balancing hyperparameters:
\begin{equation}
    \small
    \mathcal{L}_{Total} = \mathcal{L}_{focal} + \lambda_{1}\mathcal{L}_{align} + \lambda_{2}\mathcal{L}_{cal} + \lambda_{3}\mathcal{L}_{gen}
\end{equation}
The CFA is an interlocked optimization system designed specifically for the medical trilemma. In severe long-tailed distributions, forced alignment alone collapses into majority class overfitting. By coupling the objectives, we establish a dynamic gradient modulation:
\begin{equation}
\small
\nabla \theta = \eta \left( (1 - p_t)^\gamma \nabla \mathcal{L}_{CE} + \lambda \nabla \mathcal{L}_{align} \right)
\end{equation}
The focal term dynamically scales the gradient magnitude. It suppresses gradients from easy majority samples, preventing them from overwhelming the semantic alignment. Concurrently, the alignment acts as a semantic regularizer against noisy hard examples, preventing overconfident hallucinations. Furthermore, the focal query vector $q_{focal}$ is a globally shared learnable parameter that interacts with dynamically varying instance-specific visual patches $p_i$, functioning as a universal semantic filter rather than memorizing fixed spatial locations.
\subsubsection{Imbalance-Resilient Classification ($\mathcal{L}_{focal}$)}
To address the long-tail distribution, we employ the $\alpha$-balanced Focal Loss \cite{Lin2017Focal}:
\begin{equation}
    \small
    \mathcal{L}_{focal} = -\alpha_t (1 - p_t)^\gamma \log(p_t)
\end{equation}
where $p_t$ is the probability of the true class. This term down-weights easy negatives, focusing optimization on hard, minority malignancies.

\subsubsection{Visual-Semantic Alignment ($\mathcal{L}_{align}$)}
To enforce semantic consistency, we align the global visual descriptor $v_{global}$ with the textual representation $t_{global}$ (derived from the ground-truth report) using a symmetric InfoNCE objective \cite{Wang2021Understanding}:
\begin{equation}
    \small
    \mathcal{L}_{align} = -\sum_{i \in B} \log \frac{\exp(\text{sim}(v_{global}^{(i)}, t_{global}^{(i)})/\tau)}{\sum_{j \in B} \exp(\text{sim}(v_{global}^{(i)}, t_{global}^{(j)})/\tau)}
\end{equation}
\subsubsection{Theoretical Justification for Implicit Grounding} 
Minimizing $\mathcal{L}_{align}$ requires maximizing the dot product $v_{global} \cdot t_{global}$. Since $v_{global} = \sum \alpha_k p_k$, applying the chain rule to the alignment loss yields the gradient with respect to the attention weight $\alpha_k$:
\begin{equation}
\small
\frac{\partial \mathcal{L}_{align}}{\partial \alpha_k} \propto - \frac{1}{\tau} \left( p_k^\top t_{global} \right)
\end{equation}
This proves that the attention weight for any specific spatial region monotonically increases during backpropagation if and only if its visual representation $p_k$ correlates heavily with the clinical text $t_{global}$. This provides theoretical validation that our alignment implicitly forces genuine visual grounding.

\begin{table*}[t!]
  \centering
  \caption{Performance comparison on ISIC 2019, Derm7pt, and ISIC 2024.}
  \label{tab:main_results}
  \resizebox{\textwidth}{!}{
  \begin{tabular}{l c ccc ccc ccc}
    \toprule
    \multirow{2}{*}{\textbf{Model}} & \multirow{2}{*}{\textbf{Params}} & \multicolumn{3}{c}{\textbf{ISIC 2019}} & \multicolumn{3}{c}{\textbf{Derm7pt}} & \multicolumn{3}{c}{\textbf{ISIC 2024 (OOD)}} \\
    \cmidrule(lr){3-5} \cmidrule(lr){6-8} \cmidrule(lr){9-11}
    & & \textbf{B-ACC}$\uparrow$ & \textbf{AUROC}$\uparrow$ & \textbf{ECE}$\downarrow$ & \textbf{B-ACC}$\uparrow$ & \textbf{AUROC}$\uparrow$ & \textbf{ECE}$\downarrow$ & \textbf{B-ACC}$\uparrow$ & \textbf{AUROC}$\uparrow$ & \textbf{ECE}$\downarrow$ \\
    \midrule
    \textit{Multimodal Models} & & & & & & & & & & \\
    EfficientNet-B4 \cite{tan2019efficientnet} & 19M & 81.1\% & 0.901 & 0.084 & 78.2\% & 0.899 & 0.093 & 79.0\% & 0.890 & 0.098 \\
    ResNet-34 \cite{he2016deep} & 22M & 79.5\% & 0.890 & 0.095 & 77.0\% & 0.885 & 0.100 & 78.0\% & 0.880 & 0.105 \\
    ConvNeXt-Base \cite{liu2022convnext} & 88M & 82.1\% & 0.908 & 0.080 & 78.8\% & 0.905 & 0.090 & 80.0\% & 0.900 & 0.090 \\
    CaFormer-B36 \cite{Yu2022MetaFormer} & 152M & 81.8\% & 0.905 & 0.082 & 78.5\% & 0.902 & 0.091 & 79.5\% & 0.895 & 0.092 \\
    HEALNet \cite{Yu2023HealthNet} & 155M & 82.3\% & 0.910 & 0.078 & 79.0\% & 0.908 & \underline{0.087} & 80.5\% & 0.903 & 0.087 \\
    \midrule
    \textit{Vision-Language Models} & & & & & & & & & & \\
    MedCLIP \cite{Wang2022MedCLIP} & 196M & 80.7\% & 0.895 & 0.091 & 77.9\% & 0.891 & 0.101 & 78.5\% & 0.887 & 0.102 \\
    Qwen2.5-VL-7B \cite{Bai2025Qwen} & 7B & 72.8\% & 0.840 & 0.132 & 61.2\% & 0.835 & 0.140 & 65.5\% & 0.820 & 0.143 \\
    LLaVA-Med-7B \cite{Li2024LLaVAMed} & 7B & 62.1\% & 0.784 & 0.125 & 60.5\% & 0.792 & 0.188 & 61.0\% & 0.808 & 0.149 \\
    SkinVL-MM \cite{Zeng2025SkinVL} & 7B & 79.1\% & 0.920 & \underline{0.065} & \underline{80.2\%} & \underline{0.915} & 0.072 & \underline{81.5\%} & \underline{0.910} & \underline{0.080} \\
    InternVL2.5-8B \cite{Chen2024InternVL} & 8B & 66.3\% & 0.825 & 0.150 & 58.4\% & 0.810 & 0.165 & 60.1\% & 0.805 & 0.155 \\
    SkinGPT-4 \cite{Zhou2024SkinGPT4} & 13B & \underline{82.5\%} & \underline{0.942} & 0.076 & 79.1\% & 0.910 & 0.088 & 81.0\% & 0.905 & 0.085 \\
    \midrule
    \textbf{SkinCLIP-VL (Ours)} & \textbf{7.4B} & \textbf{88.7\%} & \textbf{0.981} & \textbf{0.019} & \textbf{83.4\%} & \textbf{0.965} & \textbf{0.022} & \textbf{85.0\%} & \textbf{0.972} & \textbf{0.033} \\
    \bottomrule
  \end{tabular}
  }
  \vspace{-1.5em}
\end{table*}
\subsubsection{Calibration Regularization ($\mathcal{L}_{cal}$)}
To counteract overconfidence, we explicitly optimize for reliability using the Brier Score:
\begin{equation}
    \small
    \mathcal{L}_{cal} = \frac{1}{C} \sum_{k=1}^C (p_k - y_k)^2, \quad p = \text{Softmax}(z)
\end{equation}
Minimizing $\mathcal{L}_{cal}$ directly reduces the Expected Calibration Error (ECE).

\subsubsection{Generative Reasoning ($\mathcal{L}_{gen}$)}
We utilize the standard causal language modeling (CLM) loss. The generative formulation is strictly applied to the output textual tokens, defined as:
\begin{equation}
\small
\mathcal{L}_{gen} = - \sum_{t=1}^{T} \log P \left( w_t \mid w_{<t}, V_{proj}, T_{meta} \right)
\end{equation}
Crucially, gradients from $\mathcal{L}_{gen}$ backpropagate through LoRA to the shared projector, ensuring that the visual features $V_{proj}$ retain sufficient detail for fine-grained description.





\section{Experiments}

\subsection{Datasets and Protocols}
We validate SkinCLIP-VL on a composite benchmark of three heterogeneous dermatological datasets, enforcing strict patient-level splitting to prevent data leakage.

\begin{itemize}
    \item \textbf{ISIC 2019 \cite{Codella2019ISIC}:} Our primary benchmark for long-tailed classification, containing 25,331 dermoscopic images across 8 categories. It exhibits extreme class imbalance, serving as the testbed for our $\mathcal{L}_{focal}$. We synthesized semantic descriptions using bias-free templates for alignment training.
    \item \textbf{Derm7pt \cite{Kawahara2019Seven}:} A multimodal dataset of 1,011 cases (2,000 images). We leverage its structured ``7-point checklist" (e.g., \textit{``Blue-Whitish Veil"}) as ground-truth text to supervise $\mathcal{L}_{align}$ and $\mathcal{L}_{gen}$.
    \item \textbf{ISIC 2024 \cite{Rotemberg2021Patient}:} Used exclusively as an Out-Of-Distribution (OOD) test set to evaluate zero-shot generalization.
\end{itemize}

\subsection{Evaluation Metrics and Baselines}
We employ three standard metrics to assess classification and reliability:
\begin{itemize}
    \item \textbf{Balanced Accuracy (B-ACC):} The macro-average of recall ($R_c$) across $C$ classes, robust to imbalance.
    \item \textbf{AUROC:} The area under the Receiver Operating Characteristic curve, defined by True Positive Rate (TPR) and False Positive Rate (FPR).
    \item \textbf{Expected Calibration Error (ECE) \cite{Guo2017Calibration}:} We partition $N$ samples into $M=15$ bins ($B_m$) to measure the weighted gap between accuracy and confidence:
    \begin{equation}
        \small
        \text{ECE} = \sum_{m=1}^{M} \frac{|B_m|}{N} \Big| \text{acc}(B_m) - \text{conf}(B_m) \Big|
    \end{equation}
\end{itemize}

We benchmark against: 1) \textbf{Image Backbones:} EfficientNet-B4 \cite{tan2019efficientnet}, ResNet-34 \cite{he2016deep}, ConvNeXt \cite{liu2022convnext}, HEALNet \cite{Yu2023HealthNet}; and 2) \textbf{Medical VLMs:} MedCLIP \cite{Wang2022MedCLIP}, SkinGPT-4 \cite{Zhou2024SkinGPT4}, and adapted general VLMs (Qwen2.5-VL \cite{Bai2025Qwen}, InternVL2.5 \cite{Chen2024InternVL}).

\subsection{Implementation Details}
We adopt a frozen perception, adaptive reasoning paradigm. The visual encoder is a frozen CLIP \texttt{ViT-L/14} \cite{Radford2021Learning}. The decoder is initialized from \texttt{Qwen2.5-VL-7B-Instruct}. We apply LoRA ($r=64, \alpha=16$) to attention projections, reducing trainable parameters to 4.3B. Optimization uses AdamW ($lr=1e-4$) for 20 epochs on a single NVIDIA A100 (80GB).

\subsection{Quantitative Performance Analysis}
As presented in Table \ref{tab:main_results}, SkinCLIP-VL establishes a new SOTA across all benchmarks. On the primary ISIC 2019 dataset, our method achieves 88.7\% B-ACC and 0.981 AUROC, outperforming the strongest generative baseline, SkinGPT-4, by significant margins (+6.2\% B-ACC). We highlight three critical observations regarding multimodal synergy and adaptation:
\begin{enumerate}
    \item \textbf{Effective Foundation Model Adaptation:} Direct application of general-purpose VLMs yields suboptimal results (Qwen2.5-VL: 72.8\%, LLaVA-Med: 62.1\%). In contrast, SkinCLIP-VL achieves a +15.9\% gain over Qwen2.5-VL, validating that our parameter-efficient adaptation strategy effectively bridges the domain gap between natural and medical imagery.
    
    \item \textbf{Generative vs. Retrieval:} SkinCLIP-VL significantly surpasses the MedCLIP (88.7\% vs 80.7\%). This confirms that generative modeling, when constrained by our CFA loss, captures fine-grained morphological details better than global contrastive embeddings.
    
    \item \textbf{OOD Generalization:} On the unseen ISIC 2024 dataset, our model maintains high robustness (85.0\% B-ACC, 0.972 AUROC), whereas SkinGPT-4 drops to 81.0\%. This suggests that the frozen CLIP encoder successfully preserves generalizable visual priors against domain shifts.
\end{enumerate}

\begin{figure}[t]
  \centering
   \includegraphics[width=.95\linewidth]{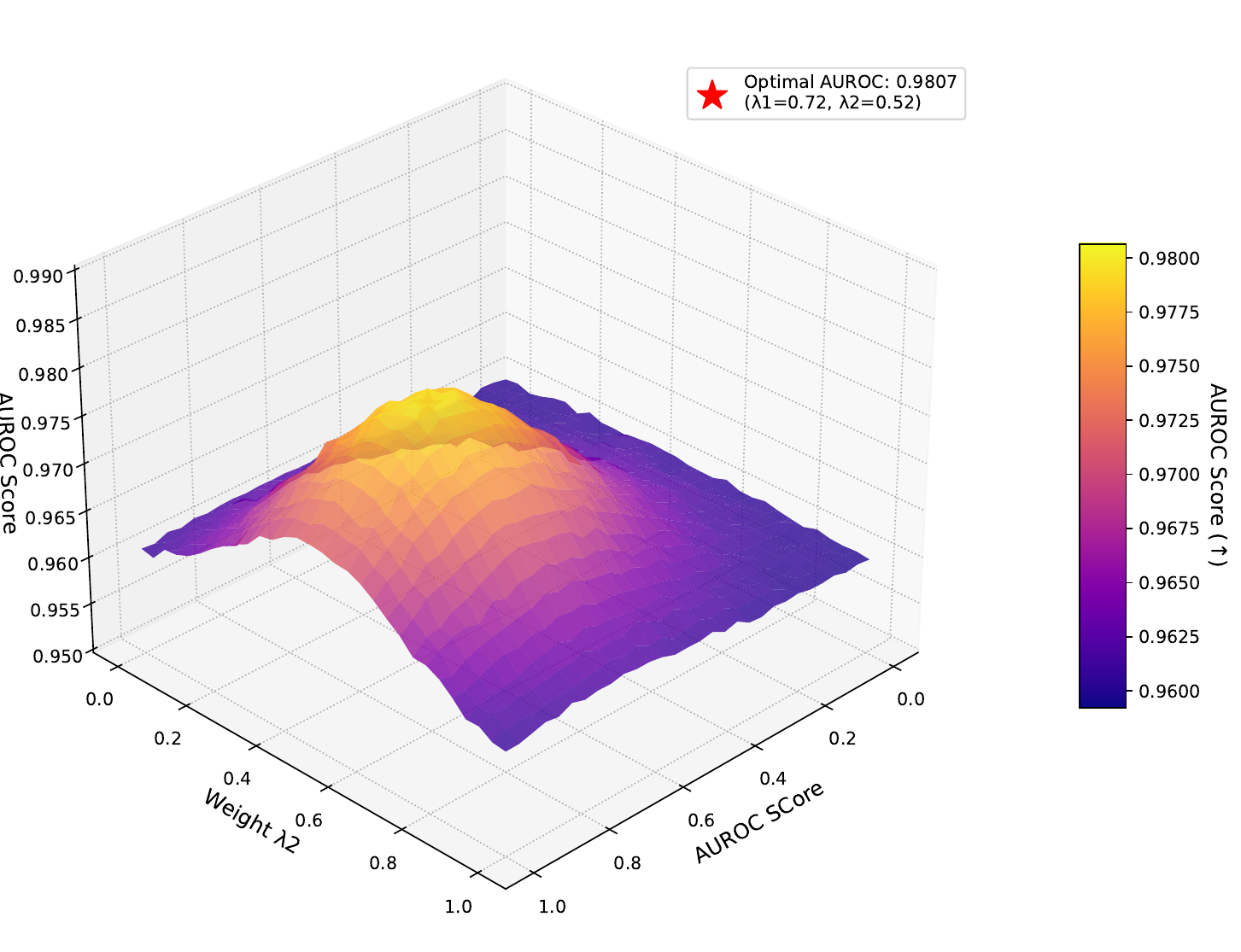}
   \vspace{-1em}
   \caption{Hyperparameter sensitivity analysis.}
   \label{fig:sensitivity}
   \vspace{-1.5em}
\end{figure}

\subsection{Reliability and System Efficiency}
High accuracy often comes at the cost of overconfidence in deep models. Notably, SkinCLIP-VL achieves an ECE of 0.019, a 75\% reduction compared to SkinGPT-4 (0.076). This demonstrates that the $\mathcal{L}_{cal}$ component effectively aligns predicted probabilities with true correctness, a prerequisite for clinical safety. Tailored for healthcare multimedia systems, our framework minimizes deployment costs. By freezing the 300M+ visual backbone and utilizing LoRA, we reduce training memory requirements by 43\% compared to full fine-tuning. This efficiency enables deployment on mid-range medical workstations, lowering the barrier for clinical adoption.

\subsection{Ablation Study}
We analyze the contribution of each component in the Consistency-aware Focal Alignment (CFA) objective using the OOD ISIC 2024 dataset (Table \ref{tab:ablation}). The CE-only baseline struggles with class imbalance (71.5\% B-ACC). Adding $\mathcal{L}_{focal}$ yields a massive +7.7\% gain, confirming that addressing the long-tail distribution is the primary driver for generalization. While $\mathcal{L}_{align}$ further boosts discrimination (+3.3\% B-ACC) by injecting semantic priors, the full CFA configuration achieves the optimal balance. Notably, it reduces the Expected Calibration Error (ECE) to 0.033 (a $\approx$79\% reduction vs. Baseline), proving that reliability can be optimized alongside accuracy.

\begin{table}[t!]
\centering
\small
\caption{Ablation of CFA components on \textbf{ISIC 2024} (OOD).}
\label{tab:ablation}
\begin{tabular}{lccc}
\toprule
Model Configuration & B-ACC $\uparrow$ & AUROC $\uparrow$ & ECE $\downarrow$ \\
\midrule
Baseline (CE Loss) & 71.5\% & 0.820 & 0.155 \\
+ $\mathcal{L}_{focal}$ & 79.2\% & 0.910 & 0.095 \\
+ $\mathcal{L}_{focal}$ + $\mathcal{L}_{align}$ & 82.5\% & 0.945 & 0.088 \\
+ $\mathcal{L}_{focal}$ + $\mathcal{L}_{cal}$ & 80.1\% & 0.915 & 0.045 \\
\textbf{+ $\mathcal{L}_{CFA}$ (Full)} & \textbf{85.0\%} & \textbf{0.972} & \textbf{0.033} \\
\bottomrule
\end{tabular}
\vspace{-1em}
\end{table}

\subsection{Data-Efficiency Analysis}
Table \ref{tab:data_eff} validates the sustainability of our framework. When training data is reduced to 12\%, baselines utilizing full fine-tuning (SkinGPT-4) or trained from scratch (EfficientNet) suffer catastrophic performance drops (over 16\%). In contrast, SkinCLIP-VL retains 97.6\% of its original performance (dropping only 2.6\%). This resilience stems from the frozen perception strategy, where the fixed CLIP encoder provides robust visual priors that require minimal adaptation data.

\begin{table}[t!]
\centering
\small
\caption{Data Efficiency Analysis on \textbf{ISIC 2024}.}
\label{tab:data_eff}
\begin{tabular}{lccc}
\toprule
Model & Train-100\% & Train-12\% & $\Delta$ Drop \\
\midrule
EfficientNet-B4 & 79.0\% & 58.3\% & -20.7\% \\
SkinGPT-4 & 81.0\% & 64.1\% & -16.9\% \\
\textbf{SkinCLIP-VL} & \textbf{85.0\%} & \textbf{82.4\%} & \textbf{-2.6\%} \\
\bottomrule
\end{tabular}
\vspace{-1em}
\end{table}

\subsection{Hyperparameter Sensitivity}
We further investigate the impact of the generation loss weight $\lambda_3$ on model performance (Fig. \ref{fig:sensitivity}). The model exhibits robustness across a wide range of $\lambda_3 \in [0.1, 1]$, with optimal performance peaking at $\lambda_3 = 0.5$. This indicates that the generative objective acts as an effective regularizer without dominating the discriminative task.
\subsection{Clinician Trust and Interpretability}

To assess clinical utility, we conducted a blinded crossover study with 20 board-certified dermatologists reviewing 50 challenging cases. Compared to text-only (SkinGPT-4) and static saliency (EfficientNet) baselines, SkinCLIP-VL achieved significantly higher expert ratings for Trust (5.2/7.0) and Justification (5.3/7.0), as detailed in Table \ref{tab:trust}. 

\begin{table}[h]
\centering
\small
\vspace{-1em}
\caption{Blinded expert evaluation (N=20, 7-point Likert).}
\label{tab:trust}
\begin{tabular}{lccc}
\toprule
Metric &  SkinGPT-4 & \textbf{Ours} \\
\midrule
1. I \textbf{trust} this output. & 4.5 & \textbf{5.2} \\
2. The rationale is \textbf{clear}. & 5.4 & \textbf{5.5} \\
3. Justifies diagnosis. & 4.2 & \textbf{5.3} \\
\bottomrule
\end{tabular}
\vspace{-.5em}
\end{table}

\begin{figure}[t]
  \centering
   \includegraphics[width=\linewidth]{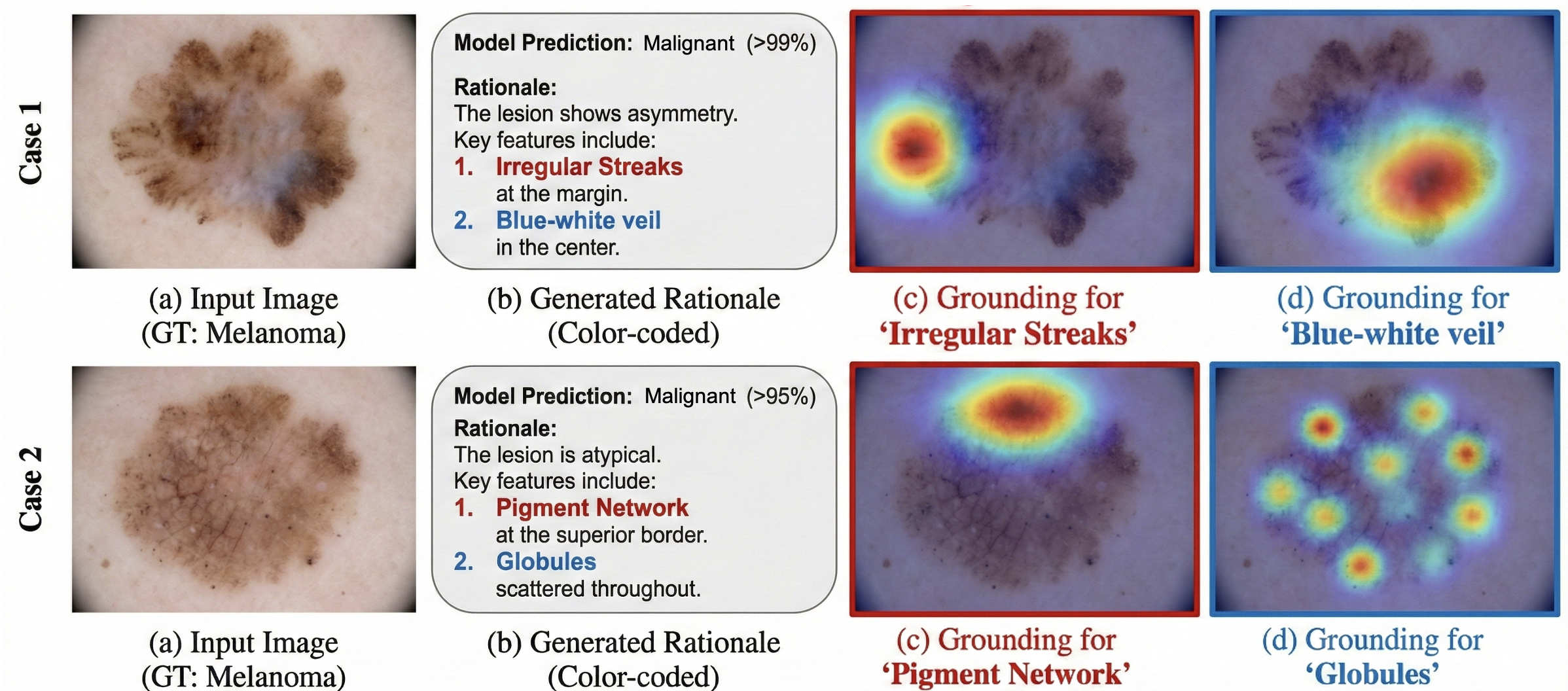}
   \caption{Case study of dynamic visual grounding.}
   \label{fig:grounding}
   \vspace{-1.5em}
\end{figure}

This clinical preference stems from our Dynamic Visual Grounding mechanism (Fig. \ref{fig:grounding}), which establishes explicit word-region correspondence, such as attending to lesion peripheries for ``irregular streaks" versus centers for ``blue-white veils". By transforming opaque predictions into verifiable, transparent reasoning, our framework effectively mitigates the ``black box" deficit and fosters actionable human-AI collaboration.

\section{Conclusion}
We presented SkinCLIP-VL, a resource-efficient framework that successfully adapts vision-language foundation models for dermatological multimedia systems. By synergizing a frozen CLIP perception module with our novel CFA loss, we effectively resolve the critical trade-offs between data efficiency, diagnostic accuracy, and predictive calibration. Extensive experiments demonstrate that our method establishes new SOTA performance on ISIC benchmarks while reducing computational overhead by 43\% and maintaining robustness under data scarcity. Ultimately, by providing clinician-verified, visually grounded rationales, SkinCLIP-VL paves the way for deploying trustworthy and accessible multimodal AI in resource-constrained healthcare environments.

\bibliographystyle{IEEEbib}
\bibliography{icme2026references}

\end{document}